%% file: main.tex
\documentclass[11pt,a4paper]{article}

\usepackage[margin=1in]{geometry}

\usepackage{indentfirst}

\usepackage[T1]{fontenc}
\usepackage{lmodern}
\usepackage{microtype}

\usepackage{amsmath}
\usepackage{amssymb}
\usepackage{amsthm}

\usepackage{graphicx}
\graphicspath{{./}}
\usepackage{float}
\usepackage{xcolor}

\usepackage{booktabs}
\usepackage{multirow}
\usepackage{array}

\usepackage{algorithm}
\usepackage{algpseudocode}

\usepackage{etoolbox}

\usepackage{enumitem}

\usepackage[colorlinks=true,linkcolor=blue,citecolor=blue,urlcolor=blue]{hyperref}
\usepackage{cleveref}

\usepackage[font=small,labelfont=bf]{caption}
\usepackage{subcaption}

\usepackage[numbers]{natbib}

\usepackage{balance}
\usepackage{comment}

\newcommand{\method}{MuJoCoUni}
\newcolumntype{L}[1]{>{\raggedright\arraybackslash}p{#1}}
\newcommand{\tabrowspace}{\addlinespace[0.42em]}

\theoremstyle{definition}

\theoremstyle{remark}

\title{%
  {\sffamily\bfseries\fontsize{22}{26}\selectfont \method{}}\\[0.4em]
  {\large Persistent Batched Runtime Primitives for MuJoCo}\\[0.3em]
  {\normalsize \url{https://github.com/unilabsim/mujoco_uni}}
}
\author{%
  Yufei Jia$^{*}$, \quad Junzhe Wu\\[0.3em]
  {\normalsize Tsinghua University}\\[0.2em]
  {\small $^{*}$Correspondence to: \texttt{jyf23@mails.tsinghua.edu.cn}}
}
\date{}

\begin{document}

\maketitle

\begin{abstract}
\input{Sections/0_Abstract}
\end{abstract}

\input{Sections/1_Introduction}
\input{Sections/2_RelatedWork}
\input{Sections/3_SystemDesign}
\input{Sections/4_Experiments}
\input{Sections/5_Applications}
\input{Sections/6_Discussion}

\bibliographystyle{plainnat}
\bibliography{references}

\end{document}

%% file: Sections/0_Abstract.tex
We present \method{}, a downstream MuJoCo distribution for online robot learning and batched physics evaluation.
Alongside the open-loop batched trajectory generation already provided by upstream \texttt{mujoco.rollout}, \method{} supplies runtime primitives for stateful environment execution.
The target workloads need high-throughput parallel execution while retaining upstream CPU MuJoCo semantics for models, sensors, contact, and constraints.
Its core object, \texttt{BatchEnvPool}, is a C++/pybind11 executor that owns per-environment \texttt{mjModel} copies, per-thread \texttt{mjData} workers, and an internal thread pool.
It provides final-state-only short stepping, sparse reset, reset-lifecycle domain randomization, batched sensor forward evaluation without advancing dynamics, and batched Jacobian and height-field queries.
The implementation is confined to the Python binding layer; MuJoCo's solver, contact model, integrator, and core source tree retain upstream semantics.
This report describes the \texttt{BatchEnvPool} API, implementation boundary, relationship to \texttt{rollout}, and the validation and benchmark scripts shipped with the open-source \texttt{mujoco-uni} package, which is installed with \texttt{pip install mujoco-uni}.

%% file: Sections/1_Introduction.tex
\section{Introduction}
\label{sec:introduction}

Robot-learning systems increasingly place the physics simulator inside the training loop.
The runtime sends batched controls, advances a short time window, reads sensors and task state, and resets only terminated environments.
MuJoCo already provides mature XML/MJB assets, sensors, contact solving, and debugging tools; when these fine-grained operations run at high frequency, interface overhead, object lifetime, and output shape directly affect training efficiency.

GPU-resident simulators and GPU-oriented MuJoCo backends are important paths for efficient training.
When a task also needs upstream CPU MuJoCo behavior for models, sensors, contact or constraint handling, or debugging, a CPU-side batched runtime provides a complementary route.

Upstream MuJoCo already provides batched stepping through the official \texttt{mujoco.rollout} interface.
It uses a C++ thread pool to run open-loop \texttt{mj\_step} from many initial states and returns full state and sensor trajectories.
It is appropriate for planning, system identification, trajectory optimization, and other tasks that consume full trajectory tensors.
Importantly, the persistence in \texttt{rollout} is limited to optional thread-pool reuse; environment models, data, state updates, reset semantics, and randomization lifecycles remain external to the call.

Online robot RL also needs an environment-runtime interface.
The runtime should preserve environments and model variants across calls, return only the final state after short stepping windows, and apply sparse reset with domain randomization for terminated environments.
Observation and control computation further need batched sensor forward passes, site Jacobians, and local terrain-height queries without advancing dynamics.

\method{} is a lightweight downstream distribution of MuJoCo with additions concentrated in the Python binding layer.
Its core object, \texttt{BatchEnvPool}, creates per-environment \texttt{mjModel} copies, per-thread \texttt{mjData} workers, and an internal thread pool.
It exposes \texttt{step}, \texttt{forward}, \texttt{reset}, \texttt{compute\_site\_jacobians}, and \texttt{sample\_hfield\_height}; MuJoCo's physics kernel and solver are unchanged.

The contribution of this report is engineering-oriented.
We describe the complementary relationship between \method{} and upstream \texttt{rollout}, present the persistent environment pool and reset/forward/query primitives, and summarize the repository scripts for numerical parity, field-patching tests, and micro-benchmarks.

The remainder of the report is organized as follows.
\cref{sec:related-work} discusses related batching and environment-runtime systems.
\cref{sec:methodology} describes the design and API of \texttt{BatchEnvPool}.
\cref{sec:experiments} documents validation and benchmark results.
\cref{sec:applications} summarizes target applications.
\cref{sec:discussion} discusses runtime boundaries and broader system context.

%% file: Sections/2_RelatedWork.tex
\section{Related Work}
\label{sec:related-work}

\subsection{Upstream MuJoCo Batching}

The closest interface to \method{} is MuJoCo's upstream \texttt{mujoco.rollout} module \cite{todorov2012mujoco}.
\texttt{rollout} generates open-loop trajectories from a batch of initial states and control sequences, supports single-threaded or thread-pool execution, and returns state and sensor arrays with shape \texttt{nbatch}~$\times$~\texttt{nstep}~$\times$~\texttt{dim}.
The center of the \texttt{rollout} abstraction is ``generate a full trajectory from input tensors''; the center of the \method{} abstraction is ``maintain a repeatedly interactive environment pool.''

This distinction determines their use cases.
\texttt{rollout} fits full-trajectory tasks such as planning, system identification, and trajectory optimization.
\texttt{BatchEnvPool} is complementary when tasks need per-environment models to persist across calls, short steps to return only final states, sparse reset-time patches, or batched current-state queries.

\subsection{Vectorized Environment Runtimes}

Vectorized environment runtimes organize many environments behind one interface and are a common engineering layer in RL systems.
EnvPool \cite{weng2022envpool} demonstrates the value of moving environment execution into a high-performance C++ runtime, and robot benchmarks such as ManiSkill \cite{taomaniskill3} expose batched task interfaces.
MotrixSim \cite{jia2026gs} shows a systems route that combines CPU-parallel simulation with reinforcement-learning algorithms for robot policy training.
\method{} occupies a lower-level position: rather than defining task semantics or a full learning runtime, it extends the MuJoCo binding layer for systems that need the standard \texttt{mjModel} workflow, persistent model pools, reset-time domain randomization, and batched physics queries.

\subsection{GPU-Resident Physics}

Brax \cite{freeman2021brax} implements a vectorizable and differentiable physics kernel in JAX; MJX \cite{mujoco_mjx} maps a subset of MuJoCo to JAX; Isaac Gym \cite{makoviychuk2021isaac} and Isaac Lab \cite{mittal2025isaac} provide NVIDIA GPU-resident simulation through PhysX; Genesis \cite{Genesis} and MuJoCo Warp \cite{mujoco_warp} also target GPU-side physics execution.
These systems can provide high throughput at large parallel scales, but GPU paths typically require models, contact and constraint handling, and data layout to fit an accelerator-friendly execution model.
They may therefore introduce constraints in feature coverage, static-shape assumptions, contact/constraint buffers, or backend hardware.

\method{} takes a complementary route.
It preserves MuJoCo CPU physics semantics and concentrates batched execution plus common robot-task queries in the C++ binding layer.
It is not a replacement claim against GPU-resident simulation; it is a CPU-batched backend for MuJoCo workloads where feature coverage matters more than accelerator residency.

\subsection{Domain Randomization}

Domain randomization is a basic technique for sim-to-real training and robust policy search.
Standard MuJoCo Python workflows typically copy or mutate \texttt{mjModel} fields and call \texttt{mj\_setConst} when required.
\method{} moves common field patches into \texttt{BatchEnvPool.reset}, so sparse reset can handle both state reset and per-environment randomization.

\subsection{Evolutionary and Optimization Workloads}

Evolutionary computing, neuroevolution, and model search also rely on large numbers of physics evaluations.
Recent work comparing CPU MuJoCo and GPU/MJX backends reports systems tradeoffs across model complexity and population size \cite{eynaliyev2025combining}.
\method{}'s persistent model pools, model-variant initialization, and final-state return semantics fit workloads that evaluate many candidate bodies or controllers in parallel.

%% file: Sections/3_SystemDesign.tex
\section{System Design and API}
\label{sec:methodology}

\begin{figure}[t]
\centering
\includegraphics[width=\linewidth]{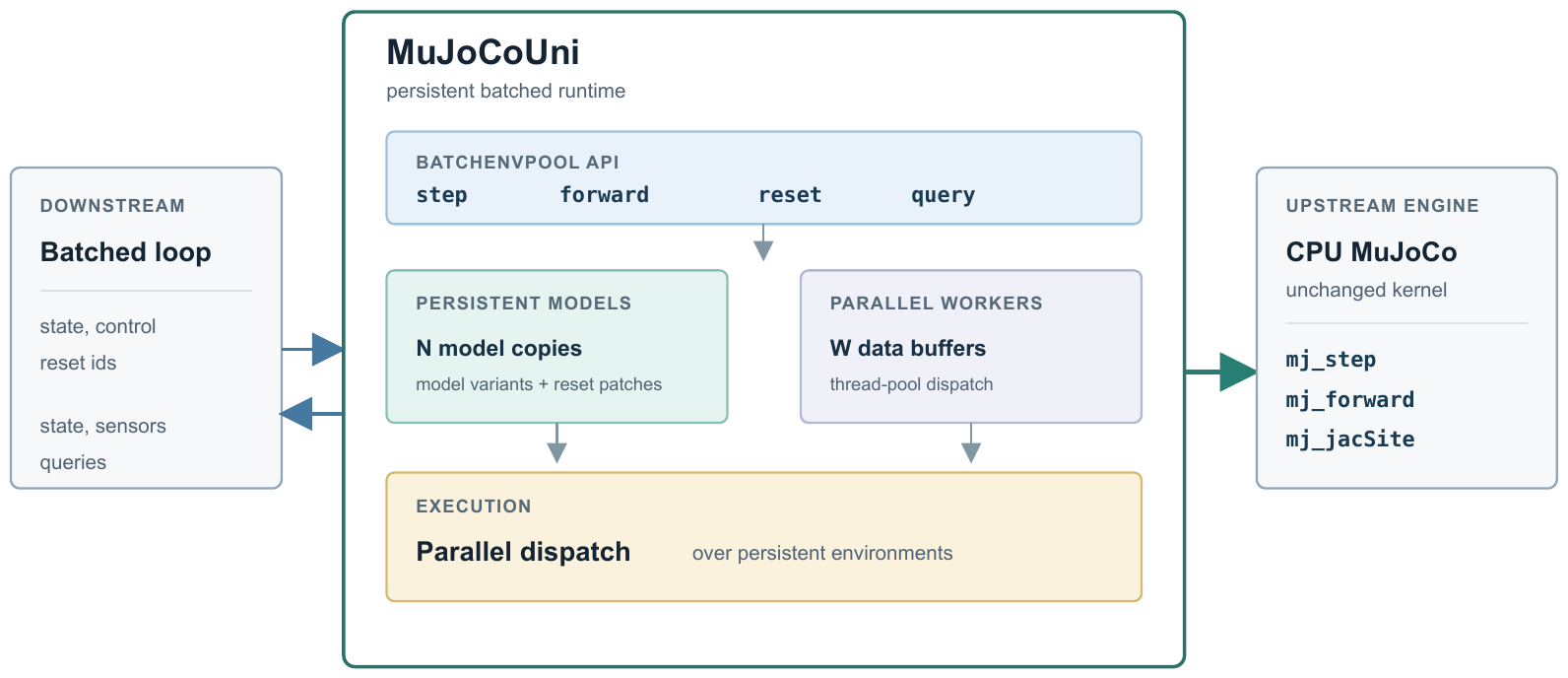}
\caption{\method{} architecture. \texttt{BatchEnvPool} maintains persistent model and worker resources behind the Python interface and executes batched operations through standard MuJoCo calls without modifying the physics kernel.}
\label{fig:architecture}
\end{figure}

\subsection{Design Boundary}
\label{sec:design-boundary}

\method{} has a narrow design boundary: it adds a batched runtime inside the MuJoCo Python package without changing the physics kernel.
Throughput improvements come from object lifetime, thread scheduling, and batched interfaces rather than from reducing the MuJoCo physics feature set.
At the code level, the core additions are \texttt{python/mujoco/batch\_env.cc} and \texttt{python/mujoco/batch\_env.py}; changes to \texttt{structs.h} and \texttt{structs\_wrappers.cc} only support safe non-owning Python references to pool-owned models.

\Cref{fig:architecture} summarizes this boundary: a downstream batched loop calls \texttt{BatchEnvPool}, which keeps per-environment models and per-worker data persistent while dispatching execution to the unchanged CPU MuJoCo kernel.

\begin{table}[t]
\centering
\small
\renewcommand{\arraystretch}{1.18}
\begin{tabular}{L{0.24\linewidth}L{0.34\linewidth}L{0.34\linewidth}}
\toprule
\textbf{Aspect} & \textbf{\texttt{mujoco.rollout}} & \textbf{\texttt{BatchEnvPool}} \\
\midrule
Primary abstraction & Stateless open-loop trajectory call; optional persistent thread pool & Stateful pool of $N$ environments owned by one executor \\
\tabrowspace
Model/data ownership & Caller supplies \texttt{mjModel}/\texttt{mjData} on each call & Pool owns per-environment \texttt{mjModel} copies and per-thread \texttt{mjData} workers \\
\tabrowspace
Step output & Full trajectory \texttt{(N, T, nstate)} and optional sensor trajectory & Final state \texttt{(N, nstate)} and optional final sensor data \\
\tabrowspace
Reset semantics & No reset primitive; caller constructs the next initial state/model externally & Sparse reset over \texttt{env\_ids} with optional field patching \\
\tabrowspace
Current-state queries & Not exposed as separate pool operations & Batched \texttt{forward}, site Jacobian, and hfield sampling \\
\tabrowspace
Best fit & Planning, system identification, trajectory optimization & Online robot RL and other stateful batched runtimes \\
\bottomrule
\end{tabular}
\caption{Interface-level distinction between upstream \texttt{rollout} and \method{}. \method{} adds persistent environment ownership and reset/query lifecycle primitives around MuJoCo's stepping semantics.}
\label{tab:rollout-comparison}
\end{table}

\subsection{Pool Construction}
\label{sec:pool-construction}

\texttt{BatchEnvPool(model, *, nbatch, nthread=None)} accepts either one \texttt{MjModel} or a compatible model sequence of length $1$ or \texttt{nbatch}.
All arguments after \texttt{model} are keyword-only.
The constructor creates one model copy per environment with \texttt{mj\_copyModel} and one \texttt{mjData} per worker thread.
When \texttt{nthread > 0}, an internal thread pool assigns chunks of environment indices to workers; otherwise execution runs on the calling thread.

This construction supports parameter-level randomization through reset-time field patches and geometry-level randomization through precompiled \texttt{MjModel} variants.
The latter covers changes such as link lengths, mesh scales, or collision geometry that cannot be expressed by simple field patches.

The pool owns its models, but downstream code sometimes needs model metadata such as sensor layout, body/site/geom ids, or playback models.
\texttt{get\_model} and \texttt{get\_all\_models} return non-owning references, avoiding double-free behavior when the pool is destroyed.

\subsection{Execution Primitives}
\label{sec:execution-primitives}

\texttt{BatchEnvPool} exposes primitives for common online robot-training operations.

\begin{table}[t]
\centering
\small
\renewcommand{\arraystretch}{1.18}
\begin{tabular}{L{0.27\linewidth}L{0.33\linewidth}L{0.32\linewidth}}
\toprule
\textbf{Primitive} & \textbf{Input pattern} & \textbf{Output / purpose} \\
\midrule
\texttt{step} & Full-pool state \texttt{(N, nstate)}, \texttt{nstep}, optional control trajectory & Final state \texttt{(N, nstate)}; optional final sensordata \\
\tabrowspace
\texttt{forward} & Full-pool state \texttt{(N, nstate)} & Sensordata \texttt{(N, nsensordata)} without advancing dynamics \\
\tabrowspace
\texttt{reset} & Subset \texttt{env\_ids}, subset states, optional randomization dict & Reset state and sensordata for only the selected environments \\
\tabrowspace
\texttt{compute\_site\_jacobians} & Full-pool state and one or more site ids & Batched translational and/or rotational site Jacobians \\
\tabrowspace
\texttt{sample\_hfield\_height} & Full-pool state, hfield geom id, local XY offsets, frame body id & Batched terrain heights or frame-to-terrain clearances \\
\bottomrule
\end{tabular}
\caption{Core \texttt{BatchEnvPool} primitives. $N$ denotes \texttt{nbatch}.}
\label{tab:api-summary}
\end{table}

\medskip
\noindent\textbf{Batched stepping.}
\texttt{step(initial\_state, nstep, control=None)} runs \texttt{mj\_step} for \texttt{nstep} integration steps on every environment.
Controls are passed as \texttt{(N, nstep, ncontrol)}, and \texttt{control\_spec} can include MuJoCo state items such as \texttt{xfrc\_applied}.
The default output is the final full-physics state.
With \texttt{return\_sensor=True}, the function also returns final-step \texttt{sensordata}; \texttt{post\_step\_forward\_sensor=True} runs one additional \texttt{mj\_forward} on the final state before copying sensors.

\medskip
\noindent\textbf{Forward evaluation.}
\texttt{forward(initial\_state)} runs one \texttt{mj\_forward} over all environments and returns sensors.
It is used to build observations from current state, refresh sensor caches after pool construction, or evaluate auxiliary signals without advancing dynamics.

\medskip
\noindent\textbf{Sparse reset.}
\texttt{reset(env\_ids, initial\_state, randomization=None)} acts only on a selected subset of environments.
\texttt{initial\_state} and every randomization payload have leading dimension \texttt{len(env\_ids)}, matching the reset subset size.
This design makes reset cost scale with the number of environments that actually terminated.

\medskip
\noindent\textbf{Site Jacobians.}
\texttt{compute\_site\_jacobians} computes \texttt{jacp} and/or \texttt{jacr} for one or more sites.
For each environment, the implementation sets state, runs the position-stage prefix \texttt{mj\_kinematics} and \texttt{mj\_comPos}, and calls \texttt{mj\_jacSite}.
For $K$ sites, outputs have shape \texttt{(N, K, 3, nv)}; when the Python wrapper receives one scalar site id, it squeezes the $K$ dimension.

\medskip
\noindent\textbf{Height-field sampling.}
\texttt{sample\_hfield\_height} bilinearly samples a MuJoCo hfield geom.
Query points are local XY offsets attached to a frame body and can be aligned in \texttt{yaw}, \texttt{world}/\texttt{none}, or \texttt{body}/\texttt{full} mode.
The output is either world-frame terrain height or clearance \texttt{frame\_z - terrain\_z}.

\subsection{Reset-Time Domain Randomization}
\label{sec:domain-randomization}

The reset randomization payload is a dictionary from field name to \texttt{float64} arrays.
Fields marked as requiring refresh trigger \texttt{mj\_setConst} for the target environments after patching; other fields are written directly before reset/forward execution.

\begin{table}[t]
\centering
\begin{minipage}{0.8\linewidth}
\centering
\small
\captionsetup{justification=centering}
\renewcommand{\arraystretch}{1.16}
\begin{tabular}{L{0.22\linewidth}L{0.18\linewidth}L{0.5\linewidth}}
\toprule
\textbf{Field} & \textbf{\texttt{mj\_setConst}} & \textbf{Use case} \\
\midrule
\texttt{body\_mass} & yes & Body mass and payload randomization \\
\tabrowspace
\texttt{body\_ipos} & yes & Inertial-frame COM offsets \\
\tabrowspace
\texttt{body\_iquat} & yes & Inertial-frame orientation perturbations \\
\tabrowspace
\texttt{body\_inertia} & yes & Inertia tensor randomization \\
\tabrowspace
\texttt{dof\_armature} & yes & Joint armature perturbations \\
\tabrowspace
\texttt{gravity} & no & Per-env gravity vectors \\
\tabrowspace
\texttt{geom\_friction} & no & Contact friction randomization \\
\tabrowspace
\texttt{kp}, \texttt{kd} & no & Position-actuator gain randomization \\
\bottomrule
\end{tabular}
\caption{Supported reset-lifecycle model patches in \texttt{BatchEnvPool}.}
\label{tab:dr-fields}
\end{minipage}
\end{table}

%% file: Sections/4_Experiments.tex
\section{Validation and Benchmarks}
\label{sec:experiments}

This section reports \method{} benchmarks on four MuJoCo models; all use the \texttt{discardvisual} compiler option to exclude visual geometries from physics computation.

\subsection{Benchmark Setup}
\label{sec:benchmark-setup}

All reported data are collected on an x86\_64 machine running Ubuntu 20.04 with an Intel i9-14900HX CPU.
The experiments use MuJoCoUni 3.8.0, Python 3.13, and NumPy 2.4, with 16 simulation threads enabled.
The C++ fast path uses 5 warmup iterations and 50 timed repetitions; Python baselines use fewer repetitions due to their significantly longer execution time.

\subsection{Correctness Coverage}
\label{sec:correctness-coverage}

\texttt{batch\_env\_test.py} covers pool construction, model sequences, input validation for \texttt{step}/\texttt{forward}/\texttt{reset}, randomization fields, indexed field access, Jacobians, and hfield sampling.

\texttt{batch\_env\_parity\_check.py} runs two parity checks: step against \texttt{rollout}, and forward against per-environment \texttt{mj\_forward} loops.

\subsection{Step and Forward Throughput}
\label{sec:throughput-benchmarks}

\begin{figure}[H]
\centering
\begin{minipage}[c]{0.39\linewidth}
\centering
\raisebox{1.0ex}{\includegraphics[width=\linewidth]{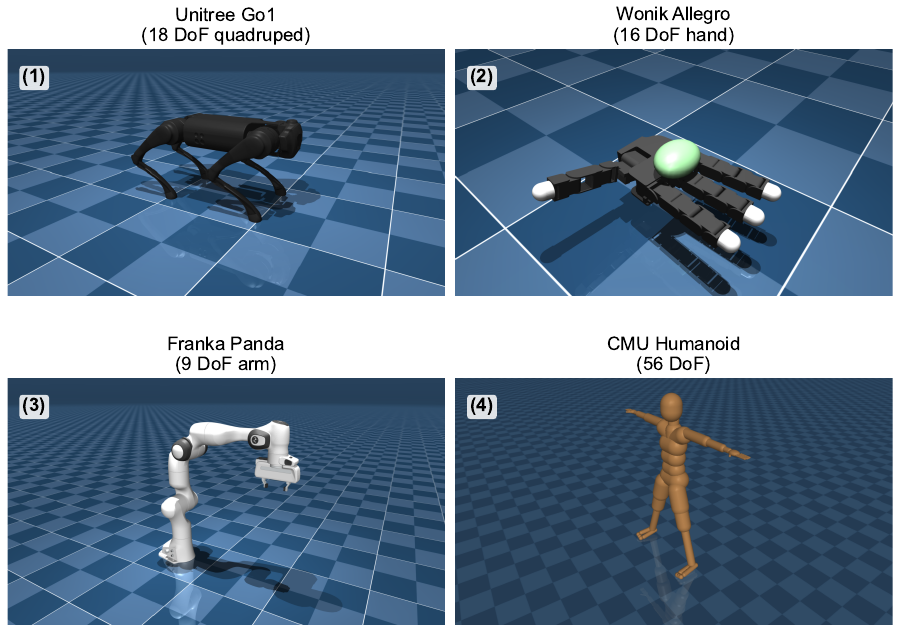}}
\end{minipage}
\hfill
\begin{minipage}[c]{0.6\linewidth}
\centering
\includegraphics[width=\linewidth]{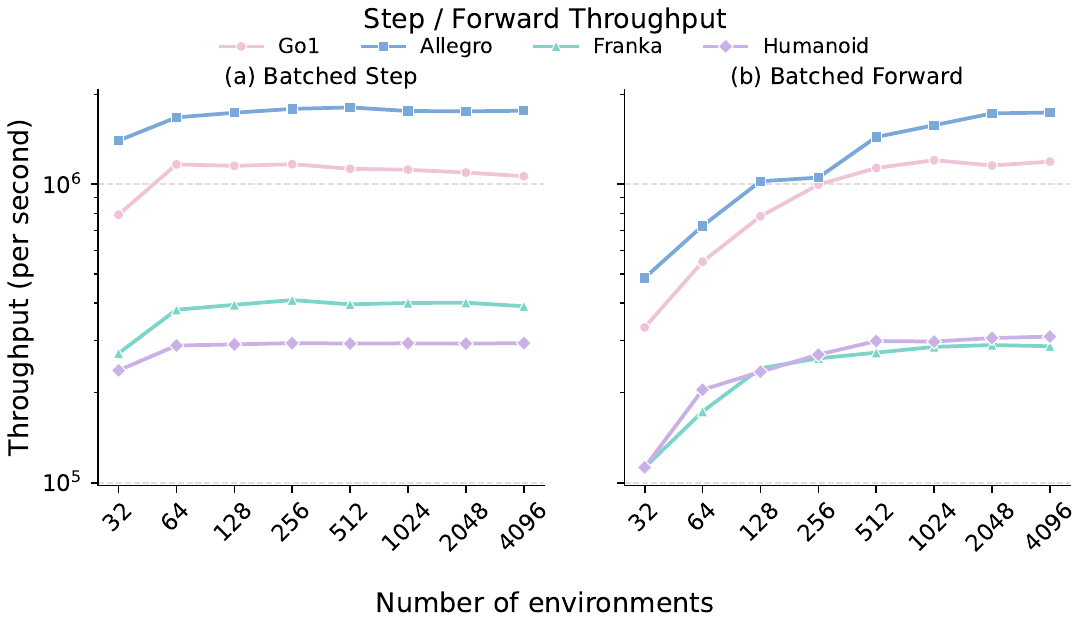}
\end{minipage}
\caption{Robot benchmark models and throughput. Left: the four robot models used in the throughput benchmarks--(1) Unitree Go1, (2) Wonik Allegro, (3) Franka Panda, and (4) CMU Humanoid. Right: batched \texttt{step} and \texttt{forward} throughput for these models; throughput saturates when the 16-thread pool becomes fully utilized.}
\label{fig:step-forward}
\end{figure}

Four models are tested: Unitree Go1 (18 DoF quadruped), Wonik Allegro (16 DoF dexterous hand), Franka Emika Panda (9 DoF arm), and a CMU Humanoid (56 DoF).
\Cref{fig:step-forward} scales the number of parallel environments from 32 to 4096; throughput saturates around 256--512 environments on the 16-thread pool.
At saturation, Allegro reaches ${\sim}1.8$M steps/s, Go1 ${\sim}1.2$M, Franka ${\sim}410$k, and Humanoid ${\sim}290$k.

\subsection{Model-Variant Overhead}
\label{sec:multi-model}

\begin{figure}[H]
\centering
\includegraphics[width=0.55\linewidth]{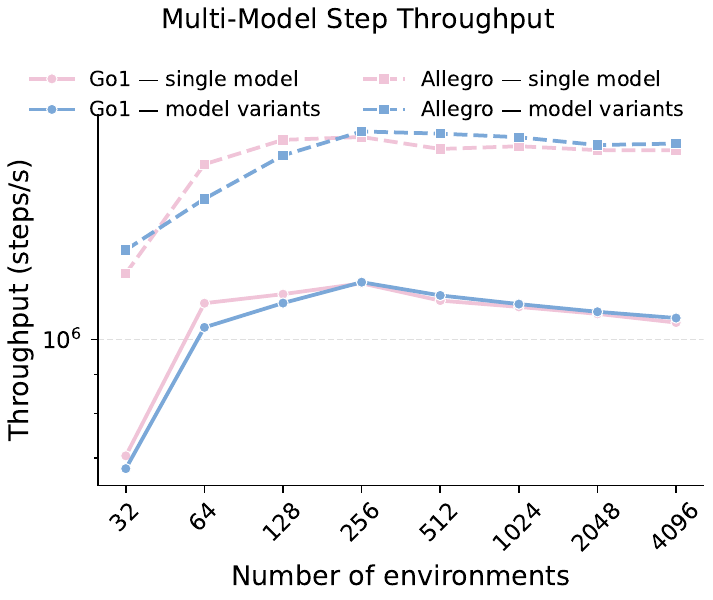}
\caption{Step throughput: single shared model versus per-environment model variants.}
\label{fig:multimodel}
\end{figure}

When each environment owns a distinct \texttt{mjModel} copy (model-variant mode), cache locality decreases slightly compared to a single shared model.
\Cref{fig:multimodel} compares the two modes for Go1 and Allegro; at saturation (256--512 environments) the gap closes and throughput is essentially identical.

\subsection{Reset Performance}
\label{sec:reset-benchmark}

\begin{figure}[H]
\centering
\includegraphics[width=0.75\linewidth]{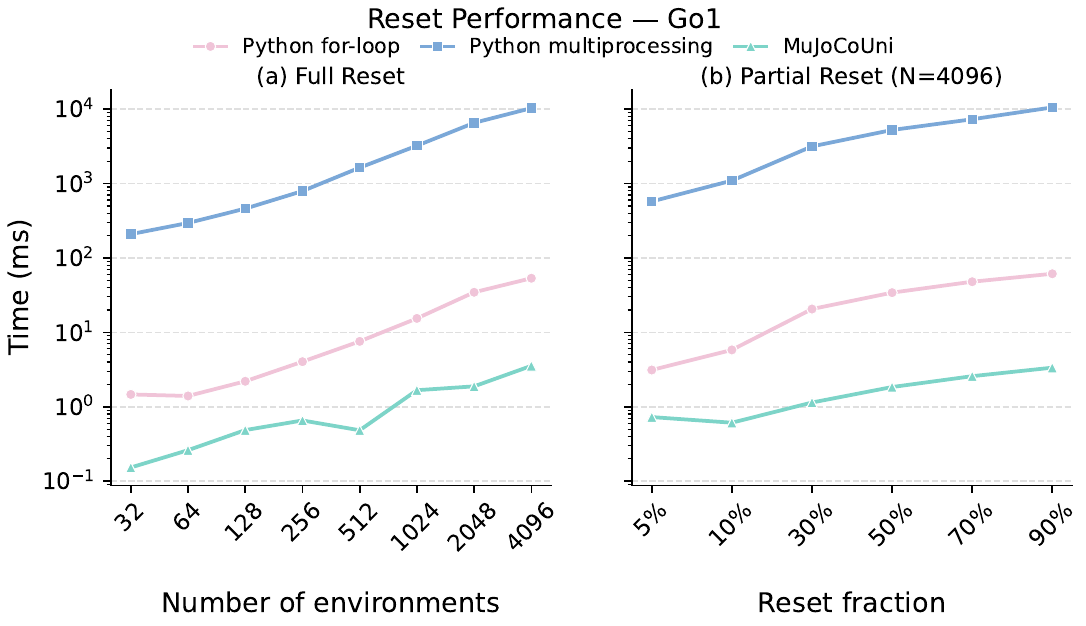}
\caption{Reset latency on Go1. Left: full reset across environment counts. Right: partial reset at 4096 environments.}
\label{fig:reset}
\end{figure}

Sparse reset is a core feature of \method{}.
\Cref{fig:reset} compares the reset latency of a Python for-loop, Python multiprocessing, and the \method{} C++ pool.
At 4096 environments, the C++ path completes a full reset in 3.5\,ms versus 53\,ms for the Python loop---a ${\sim}15\times$ speedup.
The C++ path scales linearly with reset fraction, reaching 3.4\,ms at 90\%.

\subsection{Batched Jacobian Performance}
\label{sec:jacobian-benchmark}

\begin{figure}[H]
\centering
\includegraphics[width=0.48\linewidth]{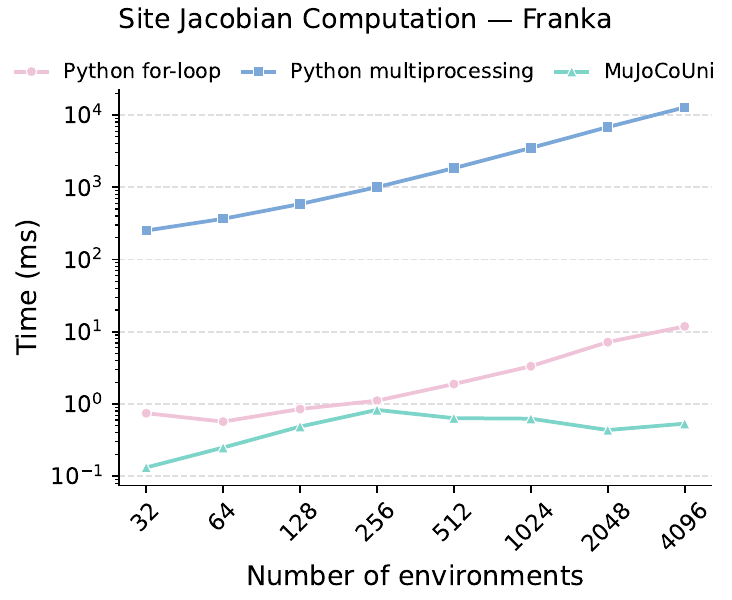}
\caption{Site-Jacobian computation time on Franka Emika Panda.}
\label{fig:jacobian}
\end{figure}

\Cref{fig:jacobian} shows the performance of \texttt{compute\_site\_jacobians} on the Franka Panda end-effector site.
The C++ pool computes Jacobians for 4096 environments in 0.53\,ms, compared to 11.9\,ms for a Python loop---a ${\sim}22\times$ speedup.

\subsection{Height-Field Sampling Performance}
\label{sec:hfield-benchmark}

\begin{figure}[H]
\centering
\begin{minipage}[c]{0.45\linewidth}
\centering
\includegraphics[width=\linewidth]{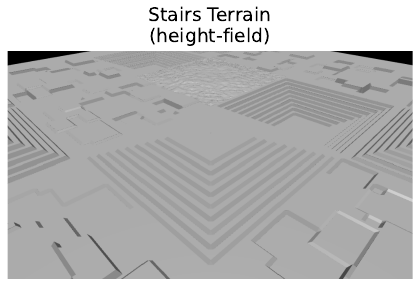}
\end{minipage}
\begin{minipage}[c]{0.48\linewidth}
\centering
\includegraphics[width=\linewidth]{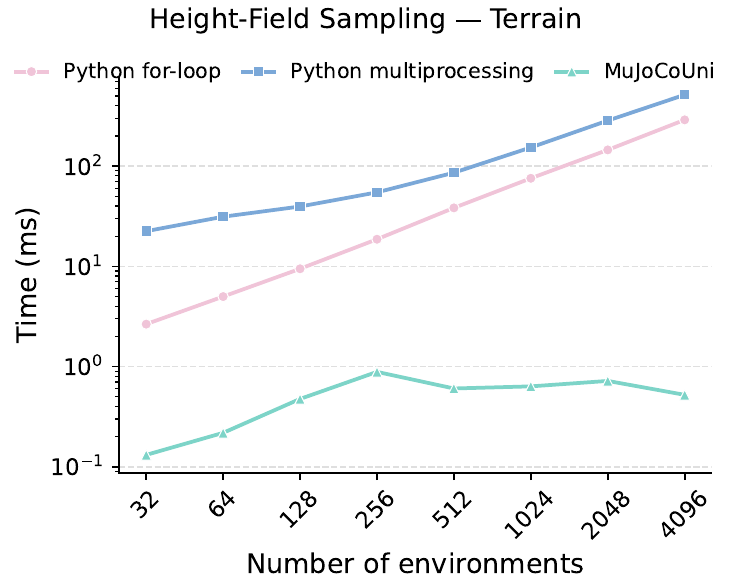}
\end{minipage}
\caption{Height-field sampling benchmark. Left: stairs height-field terrain used for batched terrain sampling. Right: height-field sampling time on the terrain model.}
\label{fig:hfield}
\end{figure}

\Cref{fig:hfield} shows \texttt{sample\_hfield\_height} performance on a stairs-terrain heightfield with a $4\times4$ sampling grid per environment.
At 4096 environments, the C++ path takes 0.52\,ms versus 290\,ms for a Python loop---a ${\sim}555\times$ speedup.

%% file: Sections/5_Applications.tex
\section{Applications}
\label{sec:applications}

\method{} provides MuJoCo-side batched runtime primitives.
It can serve as the simulation backend inside larger training, evaluation, or data-generation systems; the higher-level system remains responsible for task logic, controllers, data management, logging, and distributed scheduling.

\subsection{Robot Reinforcement Learning}
Robot RL is the primary target workload for \method{}.
\texttt{BatchEnvPool} gathers short-horizon stepping, sensor reads, sparse reset, reset-time domain randomization, and current-state queries into one MuJoCo-side object.

This interface fits several environment organizations.
Downstream systems can consume final states and sensors through synchronous batch sampling, asynchronous collection, or offline data generation.
\method{} remains neutral to controllers, learning algorithms, and data paths.

\subsection{Sim-to-Real Domain Randomization}
Sim-to-real training often randomizes dynamics parameters, contact parameters, gravity, and actuator gains during the reset lifecycle.
\method{} places common MuJoCo field patches and required \texttt{mj\_setConst} refreshes inside \texttt{reset}; geometry-level changes are represented by precompiled model variants at construction time.

\subsection{Terrain-Aware Locomotion}
Uneven-terrain locomotion tasks often require height scans in a robot-local frame.
\texttt{sample\_hfield\_height} samples MuJoCo hfield data in batch, supports yaw/world/body alignment, and returns either terrain height or frame clearance.

\subsection{Manipulation and Kinematic Control}
Manipulation, mobile manipulation, and IK tasks often need site Jacobians for end effectors or intermediate links.
\texttt{compute\_site\_jacobians} runs the minimal kinematic prefix over the full pool and calls \texttt{mj\_jacSite} in batch, supporting operational-space control, reward computation, constraint checks, IK auxiliary objectives, and analysis.

\subsection{Batch Optimization}
Evolution strategies, neuroevolution, and model/controller search often evaluate many candidates in parallel.
\method{}'s persistent model pools, model-variant initialization, and final-state-only \texttt{step} fit evaluation loops whose objective depends on final state, terminal events, or aggregated rewards.

When an optimizer needs full state and sensor trajectories at every time step, \texttt{mujoco.rollout} provides the direct full-trajectory interface.
When the evaluation loop is closer to a stateful environment pool, \method{} provides the more natural runtime model.

%% file: Sections/6_Discussion.tex
\section{Discussion}
\label{sec:discussion}

\subsection{Runtime Boundary and Tradeoffs}
\label{sec:runtime-boundary}

This report positions \method{} as a MuJoCo-compatible CPU batched runtime layer.
It places persistent environment pools, sparse reset, reset-time domain randomization, and batched sensor, Jacobian, and hfield queries in the MuJoCo binding layer while keeping physical semantics in upstream MuJoCo.

This boundary gives the main engineering tradeoffs.
Per-environment model copies increase memory use, geometry-level randomization requires precompiled compatible models, and reset-time field patching covers the currently registered field set.
The corresponding benefits are clear object ownership, lower-frequency Python interaction, and simulator-side interfaces that can be embedded in different training, evaluation, or data-generation systems.

Upstream \texttt{mujoco.rollout} covers open-loop batched trajectory generation; GPU-resident backends cover accelerator-native massive parallel simulation; \method{} covers the stateful runtime position where environments and models persist across calls.

\subsection{System Context}
\label{sec:discussion-system-context}

Efficient training stacks for modern robot-control RL are commonly organized around GPU-resident simulation and the CUDA software stack.
Systems such as Isaac Gym, Isaac Lab, MuJoCo Playground, MJLab-style MuJoCo-Warp stacks, ManiSkill, and Genesis demonstrate the practical training efficiency of large-scale GPU-parallel environments \cite{makoviychuk2021isaac, mittal2025isaac, zakka2025mujoco, mujoco_warp, taomaniskill3, Genesis}.

GPU-resident physics also has explicit implementation conditions.
Robot simulation brings dynamic contacts, sparse constraints, closed chains, and contact-rich manipulation with branch-heavy control flow and data dependencies; GPU backends therefore make tradeoffs around static allocation, fixed buffers, feature coverage, numerical formats, and solver approximations.

CPU MuJoCo provides a different set of systems properties.
It runs across platforms, preserves mature XML/MJB assets, sensors, debugging, and visualization workflows, and naturally supports physics computations with branching and irregular control flow.
Combined with modern multicore CPUs, thread pools, persistent object lifetimes, and batched array interfaces, it can serve as a high-throughput data-generation component; MotrixSim \cite{jia2026gs} has already demonstrated this route.

This interface decouples physics simulation from other components in a larger system.
Downstream systems can organize learning, logging, scheduling, rendering, or distributed execution according to their own needs while keeping the simulator side on portable CPU MuJoCo semantics.
For workloads that need full MuJoCo feature coverage, cross-platform deployment, or reuse of existing MuJoCo assets, \method{} provides a concrete CPU-batched engineering path.

\subsection{Availability and Reproducibility}
\label{sec:availability}

\method{} is released as the open-source \texttt{mujoco-uni} Python package together with unit tests and parity checks.
The benchmark code is maintained separately at \url{https://github.com/unilabsim/mujoco_uni_bench}.